\documentclass[11pt,letterpaper]{article}
\usepackage{ijcnlp2017}
\usepackage{times}
\usepackage{latexsym}
\usepackage{lingmacros}
\usepackage{graphicx}

\ijcnlpfinalcopy

\def\ijcnlppaperid{***}

\title{Complex Word Identification: \\Challenges in Data Annotation and System Performance}

\author{Marcos Zampieri\textsuperscript{1}, Shervin Malmasi\textsuperscript{2}, Gustavo Paetzold\textsuperscript{3}, Lucia Specia\textsuperscript{3} \\
\textsuperscript{1}University of Wolverhampton, United Kingdom\\
\textsuperscript{2}Harvard Medical School, United States\\
\textsuperscript{3}University of Sheffield, United Kingdom\\
  {\tt m.zampieri@wlv.ac.uk, smalmasi@bwh.harvard.edu} \\ 
{\tt g.h.paetzold@sheffield.ac.uk, l.specia@sheffield.ac.uk}}

\date{}

\begin{document}

\maketitle

\begin{abstract}
This paper revisits the problem of complex word identification (CWI) following up the SemEval CWI shared task. We use ensemble classifiers to investigate how well computational methods can discriminate between complex and non-complex words. Furthermore, we analyze the classification performance to understand what makes lexical complexity challenging. Our findings show that most systems performed poorly on the SemEval CWI dataset, and one of the reasons for that is the way in which human annotation was performed.
\end{abstract}

\section{Introduction}

Lexical complexity plays a crucial role in reading comprehension. Several NLP systems have been developed to simplify texts to second language learners \cite{petersen2007text} and to native speakers with low literacy levels \cite{specia2010translating} and reading disabilities \cite{rello2013simplify}. Identifying which words are likely to be considered complex by a given target population is an important task in many text simplification pipelines called complex word identification (CWI). CWI has been addressed as a stand-alone task \cite{shardlow:2013:SRW} and as part of studies in lexical and text simplification \cite{paetzold2016lexical}. 

The recent SemEval 2016 Task 11 on Complex Word Identification
%\footnote{\url{http://alt.qcri.org/semeval2016/task11/}} 
-- henceforth SemEval CWI -- addressed this challenge by providing participants with a manually annotated dataset for this purpose \cite{CWI}. In the SemEval CWI dataset, words in context were tagged as complex or non-complex, that is, difficult to be understood by non-native English speakers, or not. Participating teams used this dataset to train classifiers to predict lexical complexity assigning a label 0 to non-complex words and 1 to complex ones.
%Each instance in the training set was annotated by 20 annotators, and each one in the test set by only one. This setup imitates a realistic scenario where one must attempt to predict individual needs based on the needs of a group. The instances were evenly distributed across 400 non-native English speakers. 
Below is an example instance from their dataset:

\enumsentence{A {\bf frenulum} is a small fold of tissue that secures or {\bf restricts} the {\bf motion} of a mobile organ in the body.}

\noindent The words in bold --- {\em frenulum, restricts, motion} --- have been assigned by at least one of the annotators as complex
%\footnote{The notion of complexity in CWI is not necessarily related to the terms {\em simplex} and {\em complex} in morphology.} 
and thus they were labeled as such in the training set. All words that have not been assigned by at least one annotator as complex have been labeled as non-complex.  
%The SemEval CWI task is therefore modeled as a binary text classification task at the word level. 

\begin{table*}[ht!]
\centering
\scalebox{0.84}{
    \begin{tabular}{lp{11cm}c}
    \hline
    \bf Team  & \bf Approach  & \bf System Paper \\ \hline
    
    SV000gg & System voting with threshold and machine learning-based classifiers trained on morphological, lexical, and semantic features & \citep{paetzold2016svsystems} \\
    
    TALN & Random forests of lexical, morphological, semantic \& syntactic features & \citep{ronzano-EtAl:2016:SemEval} \\
    
    UWB & Maximum Entropy classifiers trained over word occurrence counts on Wikipedia documents & \citep{konkol:2016:SemEval} \\
    
    PLUJAGH & Threshold-based methods trained on Simple Wikipedia &  \citep{wrobel:2016:SemEval} \\
    
    JUNLP & Random Forest and Naive Bayes classifiers trained over semantic, lexicon-based, morphological and syntactic features & \citep{mukherjee-EtAl:2016:SemEval} \\
    
    HMC & Decision trees trained over lexical, semantic, syntactic and psycholinguistic features & \citep{quijada-medero:2016:SemEval} \\
    
    MACSAAR & Random Forest and SVM classifiers trained over Zipfian features & \citep{zampieri-tan-vangenabith:2016:SemEval} \\
    
    Pomona & Threshold-based bagged classifiers with bootstrap re-sampling trained over word frequencies & \citep{kauchak:2016:SemEval} \\
    
    Melbourne & Weighted Random Forests trained on lexical/semantic features & \citep{brooke-uitdenbogerd-baldwin:2016:SemEval} \\
    
    IIIT & Nearest Centroid classifiers trained over semantic and morphological features & \citep{palakurthi-mamidi:2016:SemEval} \\
    \hline
    \end{tabular}
}
\caption{SemEval CWI - Systems and approaches}
\label{tab:approaches}
\end{table*}

In this paper we evaluate the dataset annotation and the performance of systems participating in the SemEval CWI task. We first estimate the theoretical upper bound performance of the task given the output of the SemEval systems. Secondly, we investigate whether human annotation correlates to the systems' performance by carefully analyzing the samples of multiple annotators. Although in the shared task complexity was modeled as a binary classification task, we pose that lexical complexity should actually be seen in a continuum spectrum. Intuitively, words that are labeled as complex more often should be easier to be predicted by CWI systems. This hypothesis is investigated in Section \ref{sec:manual}. To the best of our knowledge, no evaluation of this kind has been carried out for CWI. The most similar analyses to ours have been carried out by \newcite{malmasi2015oracle} for native language identification and by \newcite{dslrec:2016} for language variety identification.

\section{Methods and Experiments}

In this section we present the data, the methods, and an overview of the experiments we propose in this paper. The goal of the experiments is to evaluate CWI performance with respect to computational methods and the manual annotation of the dataset. For this purpose we build a plurality ensemble and an oracle classifier and subsequently analyze systems output using the manual annotation provided by the SemEval CWI organizers.

%\lucia{we need to introduce here what the experiments are, their goals - a little like in the abstract, referring to specific subsections}

\subsection{Data}
\label{sec:data}

The dataset compiled for the shared task contains a training set composed of 2,237 instances and a test set of 88,221 instances. The data was collected through on-line questionnaires in which 400 non-native English speakers were presented with several sentences and asked to select which words they did not understand the meaning of. Annotators were students and staff of various universities. The training set is composed by the judgments of 20 distinct annotators over a set of 200 sentences, while the test set is composed by the judgments made over 9,000 sentences by only one annotator. 

The 9,200 sentences were evenly distributed across the 400 annotators. In the training set, a word is considered to be complex if at least one of the 20 annotators judged them so, thus reproducing a scenario that captures one of the biggest challenges in lexical simplification: predicting the vocabulary limitations of individuals based on the overall limitations of a group. This dataset is one of the few datasets available for CWI, another example is the one by \newcite{yimammultilingual}. 
%The CWI dataset is the only public available dataset of its kind. 
%It has been used not only in the SemEval CWI task but also in a recent study on complex word identification by \newcite{davoodi2017context}.

\subsection{Systems}
\label{sec:systems}

The SemEval CWI shared task provided an opportunity to compare the performance of CWI approaches using a common dataset. It was the first and only challenge organized on the topic thus far. The task was very popular, having attracted 21 teams and 42 participating systems. In Table \ref{tab:approaches} we present the 10 highest performing approaches proposed by participants of the SemEval CWI task. % with references to the system description papers. 

%\lucia{you can save some space by adding the citation next to the approach description and removing the last column}

\subsection{Approaches}
\label{sec:method-ensemble}
We build ensemble classifiers taking the output of systems that participated in the SemEval CWI task as input. This approach is equivalent to training multiple classifiers and combining them using ensembles. Our first goal is to build high-performance classifiers using plurality voting. Our second goal is to estimate the theoretical upper bound performance given the output of the systems that participated in the SemEval CWI competition using the oracle classifier.
%, which, as previously mentioned, is the only publicly available dataset compiled for this task. 
Following \newcite{malmasi2015oracle} and \newcite{dslrec:2016} we use two approaches:

%\lucia{this is a painfully long sentence and there's a verb missing - revise. Also, why mention other classification tasks? make the message clear: is the goal to estimate the theoretical upper bound (oracle)? or to get the best possible performance (ensemble)?}

%\lucia{"propose" means it's new. This has been done before, right? so we should say 'use'/'employ'} the following combination methods:

\paragraph{Plurality Voting:} This approach selects the label with the highest number of votes, regardless of the percentage of votes it received \cite{polikar2006ensemble}. 
%This differs from a \textit{majority} vote combiner where a label must obtain over $50\%$ of the votes.

\paragraph{Oracle:} It assigns the correct label for an instance if at least one of the classifiers produces the correct label for the given data point. It serves to quantify the theoretical upper limit performance on a given dataset \cite{kuncheva2001decision}.

%\lucia{these two approaches are completely different and should be presented as such: first specify what the goals of the combination are: upper bound vs consensus(?)}

\section{Results}

%In this section we present the results obtained by the ensemble classifiers. We start with the plurality voting system, and then we present the results obtained by the oracle.

\subsection{Plurality Voting}

We first test the plurality voting ensemble using the output of all 46 entries (42 runs plus 4 baselines) submitted to the CWI task.  We also built a plurality ensemble system using only the output of the top 10 systems. Our assumption was that including systems that did not perform well in the task degrades the voting performance by introducing too much noise in the predictions.

Plurality voting results for class 1 are presented in Table \ref{tab:plurality} in terms of precision, recall, and F1 score. For comparison we also report a threshold-based baseline on word frequencies from Wikipedia \cite{CWI} and the performance of the best system in terms of f-score for class 1. The number of instances in each class is presented in the column `Samples'.

\begin{table}[!ht]
\centering
\scalebox{0.90}{
    \begin{tabular}{lccccc}
    \hline
    \bf System &\bf Class & \bf P  & \bf R & \bf F1 & \bf Samples   \\ \hline
    All & 0 & 0.98  & 0.83 & 0.90 & 84,090 \\
    All & 1 & 0.17 &  0.71 &  0.27 & 4,131 \\     
    Top 10 & 0 & 0.98  & 0.88 & 0.93 & 84,090 \\
    Top 10 & 1 & 0.21 &  0.66 &  0.32 & 4,131 \\ \hline
    Baseline & 1 & 0.08 & 0.90 & 0.15 & 4,131    \\ 
   Best & 1 & 0.29 & 0.45 & 0.35 & 4,131 \\
   \hline
%     \multicolumn{2}{l}{Average Top 10} & 0.95 &  0.82 & 0.87 & 88,221 \\  \hline
    \end{tabular}
}
\caption{Results for plurality voting}
\label{tab:plurality}
\end{table}

\noindent The results obtained show that the plurality voting system performs significantly better on class 0 (non-complex words) achieving 0.90 F1 score than on class 1 (complex words) achieving 0.27 F1 score. The majority of instances in the dataset are non-complex words and this explains the bias. For class 1, the F1 score obtained by the ensemble featuring the top 10 systems outperforms the baseline but it is outperformed by the best system by 3 percentage points. 
%The results confirm our assumption and we observe an improvement of 3 percentage points F1 score in the overall performance and 5 percentage points F1 score for class 1. 

%\lucia{Add TOP and worst participating system, as well as average system performance, as in previous table?}

% \begin{table}[ht!]
% \centering
% \scalebox{0.90}{
%     \begin{tabular}{ccccc}
%     \hline
%     \bf Class & \bf P  & \bf R & \bf F1 & \bf Samples   \\ \hline
%     0 & 0.98  & 0.88 & 0.93 & 84,090 \\
%     1 & 0.21 &  0.66 &  0.32 & 4,131 \\ \hline
%     Baseline 1 & 0.08 & 0.90 & 0.15 & 4,131 \\ \hline
%     Average & 0.95 &  0.87 & 0.90 & 88,221 \\
%     \hline
%     \end{tabular}
% }
% \caption{Results for plurality voting (top-10)}
% \label{tab:pluralitytop10}
% \end{table}

% \vspace{-2mm}

%Even though this system was built with less data (less systems in the ensemble), this confirms our assumption that the voting system performs best when relying on more accurate predictions. 
%\lucia{add top performing and average of top 10} 
%\lucia{it wasn't really trained, it's just a selection - I'd remove this sentence altogether: it is expected that you'll do better if you only use good systems}

\subsection{Optimal Ensemble and Oracle}

\begin{figure}[!ht]
\centering
\includegraphics[width=.49\textwidth]{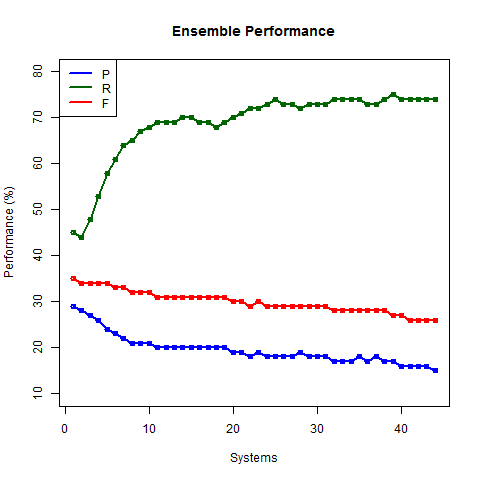}
\caption{Plurality voting using $n$ best systems}
\label{fig:optimal}
\end{figure}

We showed the performance of plurality voting ensembles built with the output of all systems and with the output of the top-10 ranked systems. The setup using the output of the top-10 systems yielded very good performance, but still below the best system in the competition. In this section we investigate how many systems should be included in the ensemble to obtain the best possible performance. In Figure \ref{fig:optimal} we show the F1-score, precision, and recall results for class 1 obtained by plurality voting using ensemble configurations ranging from 3 to 46 systems.

To investigate the optimal ensemble configuration we performed a greedy backward search over the systems, iteratively removing the worst systems in a stepwise manner without a stopping criterion. The best performance for complex words was obtained using with the predictions of the top-3 systems achieving 0.35 F1-score. This is the best performing and smallest ensemble configuration confirming that the SemEval CWI is a very challenging task which led the vast majority of systems to perform so poorly that the plurality voting ensemble did not benefit from their predictions.

\vspace{-2mm}

\vspace{-2mm}

\noindent Finally, in Table \ref{tab:oracletop3} we present the results obtained by the oracle classifier using the top-3 systems, which yielded the best results in the plurality voting ensemble. The oracle performs very well when predicting non-complex words achieving 0.98 F1-score. The performance for complex words was substantially higher than the one obtained using the configurations of the plurality voting ensemble, reaching 0.60 F1-score and outperforming both the baseline and the best system. This is the theoretical upper bound of the task given the output of the systems that used this dataset.

%\lucia{it's hard to visualise that without knowing the difference in performance between the systems. At least for the top 10, you could add P, R and F to table 1}

%\lucia{report precision and recall too for completeness - I'd add that to the plot: 2 more lines}.

%\subsection{Oracle-informed Ensemble and Upper Bound}

%\lucia{consider having one instead of 3 tables, as a lot of the figures are repeated: baseline, average, top, worst, samples ...}

\begin{table}[ht!]
\centering
\scalebox{0.90}{
    \begin{tabular}{lccccc}
    \hline
    \bf System & \bf Class & \bf P  & \bf R & \bf F1 & \bf Samples   \\ \hline
    Oracle & 0 & 0.98  & 0.98 & 0.98 & 84,090 \\
    Oracle & 1 & 0.59 &  0.61 &  0.60 & 4,131 \\ \hline
    Baseline & 1 & 0.08 & 0.90 & 0.15 & 4,131 \\
    Best & 1 & 0.29 & 0.45 & 0.35 & 4,131 \\
       \hline
%     Average & 0.96 &  0.96 & 0.96 & 88,221 \\
%     \hline
    \end{tabular}
}
\caption{Results for oracle classifier (top-3)}
\label{tab:oracletop3}
\end{table}

\vspace{-4mm}

% \begin{table}[ht!]
% \centering
% \scalebox{0.95}{
%     \begin{tabular}{ccccc}
%     \hline
%     \bf Class & \bf P  & \bf R & \bf F1 & \bf Samples   \\ \hline
%     0 & 0.99  & 0.98 & 0.99 & 84,090 \\
%     1 & 0.71 &  0.81 &  0.76 & 4,131 \\
%     \hline
%     Avg. & 0.98 &  0.98 & 0.98 & 88,221 \\
%     \hline
%     \end{tabular}
% }
% \caption{Results for oracle classifier (top-10 systems)}
% \label{tab:oracletop10}
% \end{table}

% \begin{table}[ht!]
% \centering
% \scalebox{0.95}{
%     \begin{tabular}{ccccc}
%     \hline
%     \bf Class & \bf P  & \bf R & \bf F1 & \bf Support   \\ \hline
%     0 & 1.00 & 1.00 & 1.00 & 84,090 \\
%     1 & 1.00 & 0.99 & 1.00 & 4,131 \\
%     \hline
%     Avg. & 1.00 &  1.00 & 1.00 & 88,221 \\
%     \hline
%     \end{tabular}
% }
% \caption{Results for the oracle classifier}
% \label{tab:oracle}
% \end{table}

%\lucia{I'm not sure about this term: this is the upper bound for system combination given the systems that participated in the task- it's not the upper bound for the dataset: if anyone comes up with a better system, that will raise this upper bound. So I'd rephrase this here and in previous mentions}

\subsection{Lexical Complexity}
\label{sec:manual}

\begin{figure}[!ht]
\centering
\includegraphics[width=.49\textwidth]{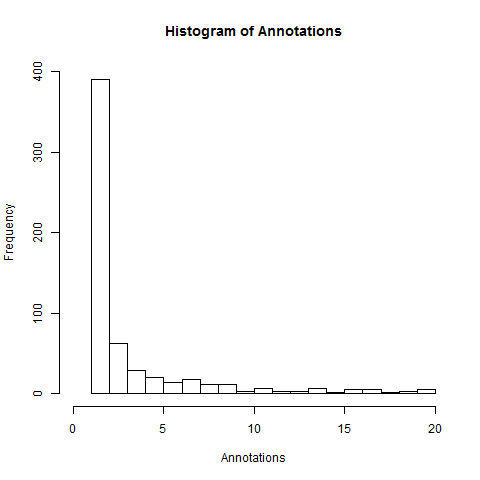}
\caption{Histogram of annotations.}
\label{fig:histogram}
\end{figure}

In this section we investigate features of the dataset and annotation that influence the output of the classifiers using the training set and the results of the 10 best performing systems.
%The assumption is that complex words tagged by the majority of the annotators as such in the training data tend to be easier for algorithms to identify. 
We start by looking at an histogram of annotations of all complex words in the training data (Figure \ref{fig:histogram}).

\vspace{-2mm}

%\vspace{-2mm}

\noindent Among the 2,237 words in the training set, 706 were labeled as complex. The histogram shows the distribution of the annotation that ranged from 393 words labeled by 1 annotator as complex and only 5 words labeled by all 20 annotators as such. 

Inspired by readability metrics \cite{kincaid1975derivation}, we looked at the average word length (AWL) of the words in the training set under the assumption that longer words tend to be more often perceived as complex. We divide the dataset in intervals according to the number of annotators that assigned each word as complex: 10-20, 1-9, and none. Results are presented in Table \ref{tab:length}.

\begin{table}[ht!]
\centering
\scalebox{0.90}{
    \begin{tabular}{cccc}
    \hline
    \bf Class & \bf Annotators & \bf Words & \bf AWL \\ \hline
    1 & 10-20 & 42 & 7.07  \\
    1 & 1-9 & 664 & 6.71  \\ \hline
    1 & 1-20 & 706 & 6.74 \\ \hline
	0 & 0 & 1,531 & 5.94 \\
    \hline
    \end{tabular}
}
\caption{Word length and complexity}
\label{tab:length}
\end{table}

\noindent We observed that words that were assigned as complex are on average longer than non-complex ones. Complex words in the dataset are on average 6.74 characters long whereas non-complex words are on average 5.94 characters long. 

Finally, we investigate the interplay between annotation and system performance by analyzing the 38 words in the training data which were labeled as complex by at least half of the annotators. We 1) check the overlap of these words in the training and test sets; 2) verify how many overlapping words received the same label in the training and test sets; 3) compute the number of times humans annotated a given word as complex (0-20) and the number of top-10 systems that labeled the word as complex (0-10). We present the scores for the words that met these criteria in Table \ref{tab:topwords}. For comparison we also present five randomly selected words labeled as complex by only one annotator which received the same label in the train and test sets. 

%We keep in mind that the test set is substantially larger than the training set and that not all words received the same label in the training and test sets. 

%\lucia{This is very interesting. I wonder if we could not do the same for various percentages: you did 50\%, but maybe we could do it also for 25\%, 75\%, 100\% (if that exists). Also: how come all random words were only marked as complex by 1 human? is that because most words were only marked as complex by only 1? the histogram I suggested would help}
%\marcos{I searched in the test for words that have been labeled as complex by just one annotator.}

%\lucia{out of how many? maybe show a simple histogram of frequency of words judged as complex by annotators (how many words judged by only 1, by 2, by 3 ... by 20)}.

%We also investigate the relationship between frequency in the British National Corpus (BNC) and the score obtained by human annotations.

\begin{table}[ht!]
\centering
\scalebox{0.90}{
    \begin{tabular}{lcc}
%    \hline
%    \multicolumn{3}{c}{\bf 10+ Complex Stats} \\ \hline
%    \multicolumn{2}{l}{\bf Total} & 38 \\
%    \multicolumn{2}{l}{\bf Overlap} & 10 \\
%   \multicolumn{2}{l}{\bf Same Label} & 5 \\ 
	\hline
    \bf Word & \bf Humans  & \bf Systems \\ 
    \hline
    gharial & 20 & 10 \\
    khachkar & 17 & 10 \\
	anoxic & 14 & 10 \\
	ubiquitous & 12 & 8 \\ 
	rebuffed & 11 & 10 \\ 
    \hline
    took & 1 & 0 \\
    better & 1 & 0\\ 
    however & 1 & 0\\
    designation & 1 & 4\\
    islands & 1 & 0\\
    \hline
    \end{tabular}
}
\caption{Annotation vs. prediction.}
\label{tab:topwords}
\end{table}

\vspace{-2mm}

%\noindent 
\noindent The CWI dataset replicates a scenario in which the vocabulary limitations of individuals is assessed based on the overall limitations of a group, 
%This makes the dataset particular in terms of train/test split and label overlap, 
as a result 50\% of the most complex words did not receive the same label in the training and test sets. Nevertheless, the results of this pilot analysis seem to confirm our hypothesis that words that were tagged more often as complex in the training set tend to be easier for CWI system to identify.

\section{Conclusion and Future Work}

This paper complements the findings from the SemEval CWI shared task report \cite{CWI} by presenting an evaluation of CWI system outputs and of the dataset used in the shared task. We were able to:  1) estimate the potential upper limit of the task considering the output of the participating systems (0.60 F1 score for complex words); 2) provide empirical evidence of the relation between word length and lexical complexity for this dataset; and 3) confirm that the performance of CWI systems in this shared task is related to non-native speakers' annotation. 

Our findings serve as a starting point for a potential re-run of the SemEval CWI task and for other studies using the 2016 dataset. In future work we would like to investigate other factors that influence lexical complexity such as word frequency and grammatical categories.

\section*{Acknowledgements}

This contribution was partially supported by the European Commission project SIMPATICO (H2020-EURO-6-2015, grant number 692819). We would like to thank the anonymous reviewers for their feedback.

\bibliography{ijcnlp2017}
\bibliographystyle{ijcnlp2017}

\end{document}